\documentclass[preprint,12pt,authoryear]{elsarticle}

\usepackage{amssymb}
\usepackage{url}
\usepackage{tipa}

\journal{Computer Speech and Language (special issue)}

\begin{document}

\begin{frontmatter}



\title{Automatic recognition of child speech for robotic applications in noisy environments}


\author[label1]{Samuel Fernando}
\author[label1]{Roger K. Moore}
\author[label2]{David Cameron}
\author[label2]{Emily C. Collins}
\author[label2]{Abigail Millings}
\author[label1]{Amanda J. Sharkey}
\author[label2]{Tony J. Prescott}

\address[label1]{Department of Computer Science, The University of Sheffield, Regent Court, 211 Portobello, Sheffield, S1 4DP, UK}
\address[label2]{Department of Psychology, The University of Sheffield, Western Bank, Sheffield, S10 2TP, UK}
\begin{abstract}
Automatic speech recognition (ASR) allows a natural and intuitive interface for robotic educational applications for children. However there are a number of challenges to overcome to allow such an interface to operate robustly in realistic settings, including the intrinsic difficulties of recognising child speech and high levels of background noise often present in classrooms. As part of the EU EASEL project we have provided several contributions to address these challenges, implementing our own ASR module for use in robotics applications. We used the latest deep neural network algorithms which provide a leap in performance over the traditional GMM approach, and apply data augmentation methods to improve robustness to noise and speaker variation. We provide a close integration between the ASR module and the rest of the dialogue system, allowing the ASR to receive in real-time the language models relevant to the current section of the dialogue, greatly improving the accuracy. We integrated our ASR module into an interactive, multimodal system using a small humanoid robot to help children learn about exercise and energy. The system was installed at a public museum event as part of a research study where 320 children (aged 3 to 14) interacted with the robot, with our ASR achieving 90\% accuracy for fluent and near-fluent speech. 
\end{abstract}


\end{frontmatter}


\section{Introduction}
\label{intro}

As part of the EU project EASEL (Expressive Agents for Symbiotic Education and Learning) we study child-robot interactions in social and educational contexts. For these interactions, the capability of automatic speech recognition (ASR) provides a natural and intuitive interface for a child to communicate with a robot. ASR provides benefits in educational applications \citep{williams00} including a more engaging and enjoyable experience for children. Additionally, such an interface provides improved learning outcomes \citep{mostow08} compared to traditional computer interfaces such as keyboards and mice.

However there are significant challenges to overcome before ASR can be used effectively in such applications. These include the intrinsic difficulty of recognising child speech due to physiological differences in the vocal tract \citep{russell07b}, clarity in pronunciation (especially for younger children), the higher intra-speaker and inter-speaker variability in children's speech \citep{gerosa07}, and the higher rate of disfluencies when compared to adult speech \citep{dejoy85}. Another problem is that robotic and computer systems are required to work well in classroom environments which can often be noisy. For both adult and child speech, high levels of background noise can be severely detrimental to ASR performance.

In this article we present a number of contributions to address these challenges. We used the Kaldi \citep{povey11} open source toolkit to develop our ASR module. This toolkit contains state of the art acoustic modelling and decoding algorithms for both the traditional Gaussian Mixture Model (GMM) approach as well as the latest deep neural network (DNN) algorithms. We evaluated both approaches in this article, training the models on existing corpora of child and adult speech. Since these data sets are relatively small, we used data augmentation approaches to improve robustness over noisy conditions and speaker variation. We also provided a close integration between the ASR module and the rest of the robotic dialogue system. This allows the module to be updated in real-time with language models relevant to the current stage of the dialogue (e.g. answer options in a multiple choice quiz). The combination of a high performance acoustic model and dynamic language model offers a robust system that works well with young children in a noisy environment, even with a relatively small amount of data used for training. 

The rest of this article has the following structure. In Section \ref{related} we briefly describe previous related work in this area. Section \ref{easel} gives an overview of the EU EASEL project and the specific use case for our ASR module. Section \ref{method} describes the development details of our ASR module, including the creation of the acoustic and language models, and how the recogniser was implemented to run in an online (live) setting and integrated with the rest of the EASEL multimodal dialogue system. Sections \ref{offline_evaluation} and \ref{live_evaluation} respectively describes offline and live experiments using the ASR module. Finally Section \ref{conclusions} concludes.

\section{Related work}
\label{related}

There are significant obstacles for developing an ASR system for robotic applications. Apart from the difficulties of recognising child speech and dealing with background noise, there are few high quality speech recognition solutions that are readily accessible to researchers who are not ASR experts. Off-the-shelf commercial applications are usually designed for specific applications such as dictation or search and are not suitable for robotic dialogue systems. There are open-source toolkits available such as Julius \citep{lee09} or HTK \citep{young97} but these can be difficult for non-experts to configure. In addition, many of these toolkits have become stagnant or underdeveloped in recent years meaning they lack the high-performance algorithms and techniques which have been developed in recent years. As a result of these challenges, there been relatively few applications of ASR for children's speech despite the desirability of such a system for many scenarios. The main applications so far have been in reading tutors \citep{mostow13} which are a very useful application but are relatively unchallenging for ASR since the text that is going to be uttered is largely known beforehand. Recently the ALIZ-E social robotics project \citep{belpaeme12} showed promising ASR results in offline experiments \citep{sommavilla14} but this system was not deployed in live interactions.

A major step forward in speech recognition technology has arisen through the emergence of the Kaldi toolkit. Table \ref{asrCompareTab} shows results from \cite{gaida14} comparing Kaldi against other major ASR toolkits, with the performance of Kaldi with the DNN approach having a third of the error rate in comparison to the other software. 

\begin{table}[htp]
\begin{center}
\begin{tabular}{|c|c|} \hline
Recognizer & WER (\%) on WSJ \\ \hline
HTK  & 19.8 \\ \hline
Julius & 23.1 \\ \hline
Sphinx & 21.4 \\ \hline
Kaldi & 6.5 \\ \hline
\end{tabular}
\caption{Comparison of ASR toolkits \citep{gaida14}, WER results for the WSJ speech corpus. The DNN approach was used in Kaldi.}
\label{asrCompareTab}
\end{center}
\end{table}

\section{The EASEL project use-case}
\label{easel}

The work on ASR described in this article is part of the wider EU EASEL project. In this project we develop robotic tutoring systems to help children learn about scientific topics. As a consortium, we study the social, psychological and educational aspects of the interactions as well as working on the technical capabilities. In this article however, we focus on our particular use-case at Sheffield, the Healthy Living Scenario. This interaction features a small humanoid robot which acts as a tutor to help children learn about healthy living, in particular about exercise and energy. 

\subsection{Physical setup for the Healthy Living Interaction}

\begin{figure}[htp]
\centering
\includegraphics[width=90mm]{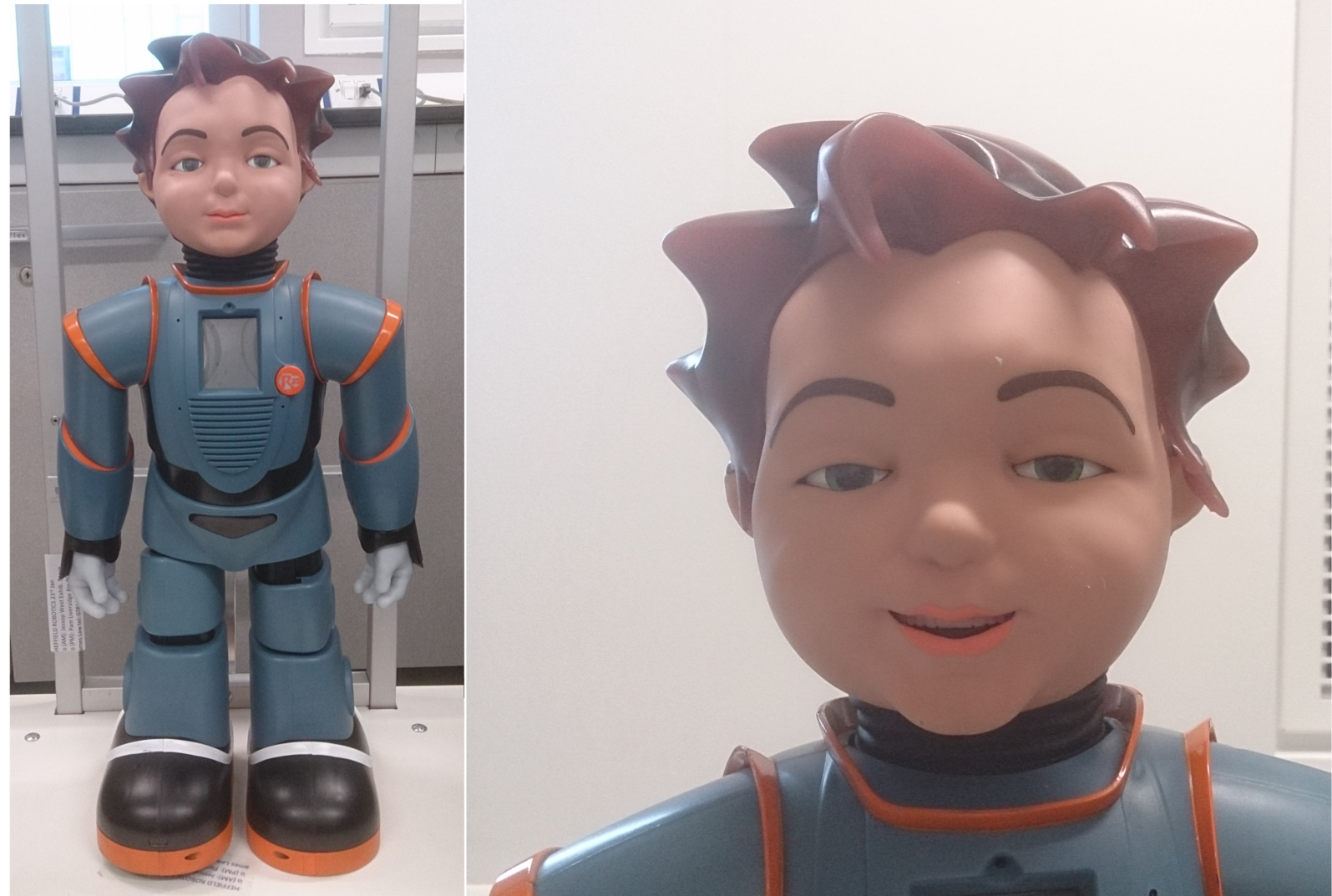}
\caption{The Zeno R25 robot, approx. 60cm tall.}
\label{zenofig}
\end{figure}

We use the Zeno R25 robot manufactured by Robokind\footnote{\url{http://www.robokindrobots.com/zeno-r25/}}, pictured in Figure \ref{zenofig}. The main feature of this robot is the emotionally expressive face with seven degrees of freedom including eyebrows, mouth opening, and smile. It also has five degrees of freedom in its arms including a grasping hand and four degrees of freedom in the legs and waist.

\begin{figure}[htp]
\centering
\includegraphics[width=130mm]{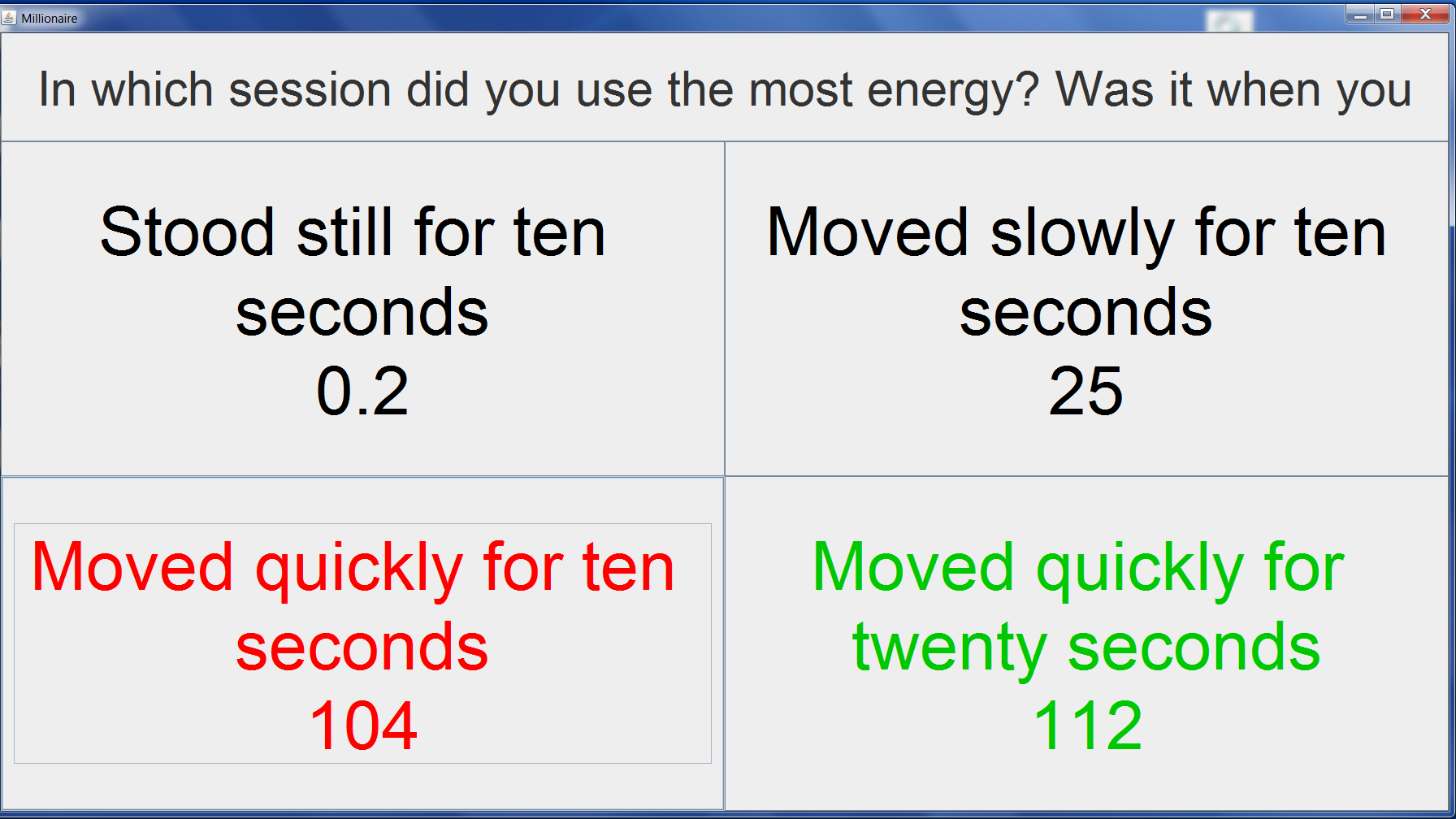}
\caption{Multiple choice question and answers displayed on screen during the Healthy Living interaction. The red colour indicates a wrong answer from the participant, with the correct answer shown in green.}
\label{quiz_screen:fig}
\end{figure}

\begin{figure}[htp]
\centering
\includegraphics[width=130mm]{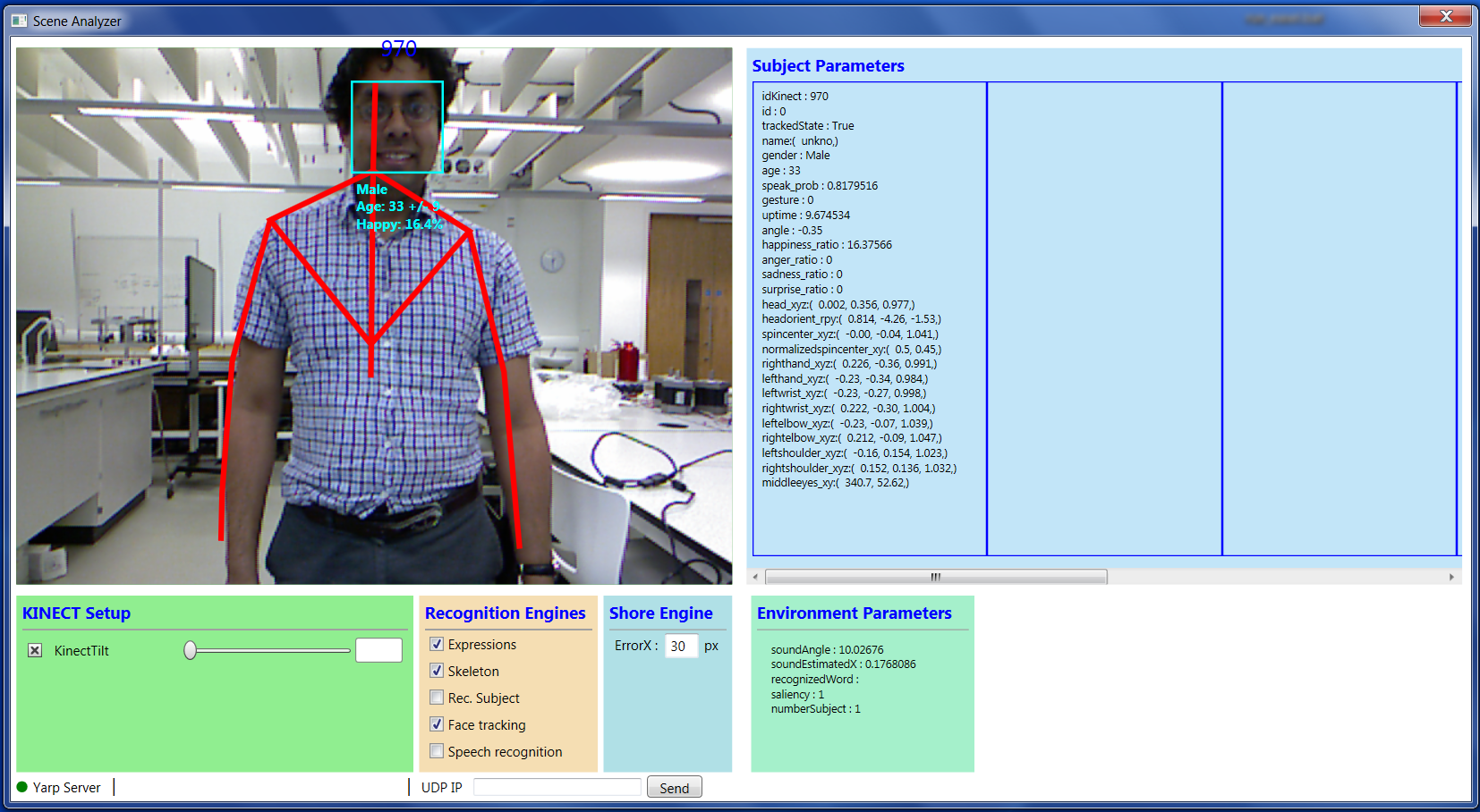}
\caption{During the exercise part of the interaction, the screen would show the skeletal tracking of the SceneAnalyser.}
\label{scene_screen:fig}
\end{figure}

The robot is equipped with its own onboard speech synthesiser based on the Acapela\footnote{\url{http://www.acapela-group.com/}} speech synthesiser. We used the default American English voice. The robot stood on a specially constructed stand. Above the stand was a mount for a Kinect sensor, used for skeletal tracking to estimate energy usage during exercise (explained in more detail in the subsequent sections). We also had a large screen TV display which was used to show prompts as well as questions and answers for a multiple choice quiz (Figure \ref{quiz_screen:fig}). During the exercise part of the interaction, the TV display would show the skeletal tracking from the SceneAnalyzer module, allowing the children to see the tracked skeleton move as they exercised, as shown in Figure \ref{scene_screen:fig}.

\begin{figure}[htp!]
\centering
\includegraphics[width=130mm]{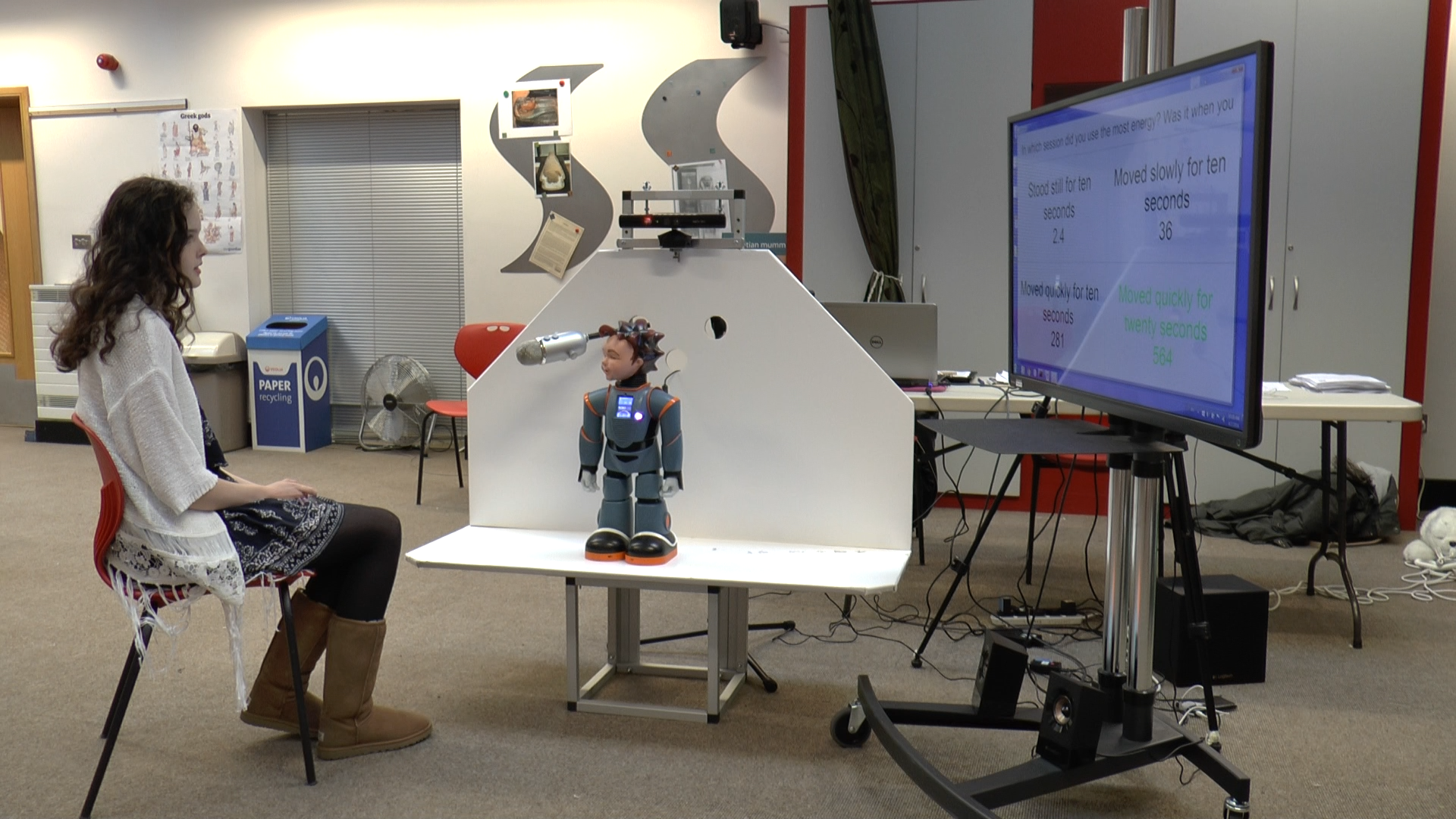}
\caption{Physical setup for the `healthy living' interaction, showing the robot, the microphone, the Kinect sensor, and the large TV screen.}
\label{system:fig}
\end{figure}

Behind the scenes, we had two laptops running all the required software for the system to run. We used the Yeti Blue microphone as our sound input for the ASR. This microphone has an USB interface so input could be processed directly without need for another sound card. A picture of the whole setup is shown in Figure \ref{system:fig}. This system was installed at the Weston Park museum in Sheffield. We occupied a large activity room which was located away from the main busy areas in the museum. To enter the room the visitors had to walk upstairs and down a corridor, and the event was clearly signposted along the way. Outside the room we put a sign saying that people were welcome to come in and participate, but to please remain quiet as speech recognition was taking place.

The system was located in one side of the room. On the other side of the room we had other activities for children to engage in, such as other robots to look at, and colouring pages for children to quietly work on. On entering the room the visitors were offered the opportunity to look at any of the events in the room. For children of around 7 or above we offered the opportunity to talk to the robot in our Healthy Living Interaction and gave a brief explanation of what the interaction was about. Younger children were not actively encouraged to participate. If they or their parents were interested we offered them the chance to participate, although we did warn that the interaction may not be suitable for younger children. For children younger than 7 we suggested that their parents or guardians may wish to sit with them and help them complete the interaction. Once a child was ready to take part, we invited them to sit in the chair, ready to talk to the robot.

For all participants we presented parents/guardians with a consent form explaining what would happen in the interaction. This was presented either as an online survey on a touchscreen tablet, or on paper. Parents and children had the opportunity to ask questions from the researchers at the event and were free to abort the interaction at any time and withdraw consent. The parents were able to tick boxes on the touchscreen or paper indicating whether they gave permission for their child to take part; gave permission for the audio to be recorded; gave permission for the video to be recorded; and finally whether they were willing to fill in a questionnaire afterwards. 

\subsection{The Healthy-Living Interaction}
\label{healthyLivingInteraction}

The first phrase spoken by the robot is: 

\begin{quote}
Robot: Hello, I am Zeeno the robot, we are going to learn about exercise and energy. First I need you to repeat some words after me so that I can get used to your voice. Please say, Hello Zeeno, I am ready to start."\footnote{Note that we used Zeeno instead of Zeno in the text on screen. This is an attempt to ensure that children pronounce this unfamiliar word in a consistent way i.e. as \textipa{[zi:n@U]} (Zeeno) instead of \textipa{[zeIn@U]} (Zeno).}
\end{quote}

The child was then expected to say: 

\begin{quote}
Child: Hello Zeeno I am ready to start.
\end{quote}

After the initial phrase the robot then asks the child to repeat two more phrases:

\begin{quote}
Robot: Thank you. Now please say, Testing A. B. C.

Child: Testing A B C.

Robot: OK great. Now please say, testing one two three.

Child: Testing one two three.
\end{quote}

These three phrases provided an initial fail-fast stage for the interaction. If the child was not able or willing to say the phrases, or if the system failed to recognise the phrases, then the researchers could abort the interaction and offer an alternative, such as looking at the other robots in the room. However if (as most often was the case) the interaction had proceeded successfully to this point, then the researchers could simply allow the interaction to continue automatically as follows:

\begin{quote}
Robot: Great thank you. Now we will do some exercise. Please step on the mat so I can see you.
 
(Child steps on mat)

Robot: OK great I can see you. Lets look at how much energy you use in exercise. I will play musical notes to show how much energy you use. If you move fast it will be high pitched, if you move slow it will be low pitched. First lets see what happens if you just stand completely still for ten seconds.
\end{quote}

The robot monitored the movement of the childs arms through the Kinect sensor and calculated the amount of energy used:

\begin{quote}
Robot: OK in that session you used a total of {\tt session1\_energy} joules of energy. Now lets see what happens when you move slowly for ten seconds. Put your arms out as far as you can. Then move them around slowly in big circles. Go!
\end{quote}

This continues two more times, with the robot varying the speed of movement requested (quickly instead of slowly), and the duration (twenty seconds instead of ten). At the end of each session the robot again reports the amount of energy used. After the fourth and final session the robot invites the child to sit down and do the quiz:

\begin{quote}
Robot: OK. You used {\tt session4\_energy} joules of energy that time. Now we will do the quiz. Please sit down on the chair and when you are ready say Zeeno Start the Quiz. 
\end{quote}

After the child said `Zeeno Start the Quiz' the quiz would start (if the phrase was not recognised then after 10 seconds the quiz would begin anyway). The quiz consisted of four multiple choice questions. The first question was:

\begin{quote}
Robot: In which session did you use the most energy? Was it when you

Stood still for ten seconds

Moved slowly for ten seconds

Moved quickly for ten seconds

*Moved quickly for twenty seconds 
\end{quote}

The robot would say the question and the four answers, and as it did so, the text would also come up on the large screen display. At the end of the last answer, the child was then expected to select and say the answer they thought was most appropriate. The next question was similar to the first except querying about when they used the least energy rather than the most (the options were the same). There then followed a general question relating to exercise and energy:

\begin{quote}
Robot: In general, which of these would use the most energy?

Watching television for twenty minutes

Reading a book for twenty minutes

Playing football for twenty minutes

Walking for twenty minutes  
\end{quote}

Again there followed a similar question about the least energy rather than the most. Finally the roles of the child and the robot were somewhat altered, in that the robot presented commands that the child could give to the robot:

\begin{quote}
Robot: Thank you for doing the quiz. Now it is your turn to tell me what to do. Choose one option. You can say

Put your left arm up

Put your right arm up

Make a happy face

Make a sad face
\end{quote}

Upon selecting an option, the robot would verbally acknowledge the command `OK I will put my left arm up' and then perform the action. Following this, another similar set of commands was presented (including an option to `Do the monkey dance' whereupon it would perform an amusing pre-programmed dance) and the robot would again do as commanded. Finally the robot concluded the interaction:

\begin{quote}
Robot: OK that is enough of that. I had fun talking with you today. Hope to see you again. Goodbye!
\end{quote}

The audio stream continued recording for a few seconds afterwards in case the child said `Goodbye' or something similar. However no recognition was performed at this point.



\section{The SF-Kaldi-ASR setup}
\label{method}

We used the Kaldi open source ASR toolkit to train our acoustic models, and develop the ASR module which we integrated into the EASEL system described above. In this section we describe the data and methods used to create this module.

\subsection{Training data}
\label{trainingData}

Our use case requires that the system to be deployed in classrooms and museums mainly in England, with British English native speakers. We therefore used corpora of British English speech to train our acoustic model. We used two corpora. The first is the British English version of the Wall Street Journal corpus created at the University of Cambridge \citep{robinson95}, which we henceforth refer to as the WSJCAM corpus. The second is the PF-star corpus of British English child speech \citep{russell06}, henceforth referred to as the PF corpus. We train with both corpora to create a single acoustic model that can be used for both adult and child speech. We used the designated portions of each corpus for training and testing.

From the WSJCAM corpus we use the training set comprising 92 training speakers, and the evaluation test set comprising 28 speakers. Each speaker provides approximately 90 utterances and an additional 18 adaptation utterances for speaker adaptation. We did not do any speaker adaptation in our experiments, and used the whole set of 108 utterances for testing. The corpus contains simultaneous recordings from both a headset microphone and a desk microphone, we use both for training and testing. We prepared our own data preparation scripts as the WSJCAM corpus differs considerably in format to the original WSJ American English corpus. However we followed the transcription normalization provided in the the American WSJ example scripts included with Kaldi. 

The PF corpus contains speech from 158 children aged 4 to 14 years. The majority of the children (excluding some of the younger children) recorded 20 SCRIBE sentences, a list of 40 isolated words, a list of 10 phonetically rich sentences, 20 generic phrases, an accent diagnostic passage (the `sailor passage') and a list of 20 digit triples. We train our system with the designated training set (86 speakers, approx 7 hrs 30 mins), and test with the evaluation test set (60 speakers, approx. 5 hrs 50 mins). This corpus contains simultaneous recordings from both a headset microphone and a desk microphone, we use both for training and testing. We prepared our own data preparation scripts to prepare the input for Kaldi. In the PF corpus there was a minor issue where `sp' and `sil' were both used to denote silence: we collapsed this down to a single symbol `!SIL' for training with Kaldi. We kept all disfluent speech. The original PF transcripts contain segmentations of individual words; we join these together for input into Kaldi as it works better with a whole utterance in each segment rather than individual words.

\subsection{Noise data for augmentation}
\label{noiseAugmentation}

In order to improve robustness in noise, we used background noise audio to augment the training data described above. For this purpose we used the CHiME corpus \citep{christensen10} which contains various kinds of background noise recorded in real-life environments. Since our main relevant use-case for the ASR is a public museum setting, we decided that the `cafe' background noise would be the best matching type of noise to use for our model. 

For each utterance in the PF and WSJCAM corpora, a section of the noisy corpus of the same length was randomly selected and added to the utterance audio. The addition was done using the SoX\footnote{\url{http://sox.sourceforge.net/}} sound processing tool, using the mix option. We added the noise at three different signal to noise levels, 5dB, 10dB and 20dB.

\subsection{Kaldi acoustic modelling}
\label{acousticModelling}
We used the Kaldi toolkit to train the acoustic models for the ASR system. The toolkit has relatively standardised scripts (collectively known as recipes) designed to work with different sets of training data. We followed the Wall Street Journal (WSJ) recipe. Training starts with a standard monophone system using standard 13 dimensional MFCCs along with first and second order derivatives. Cepstral mean normalization is used throughout to reduce the channel effect. A triphone system was then constructed using speaker-independent alignments derived from the monophone system, and a linear discriminant analysis (LDA) transform was employed to select the most discriminative dimensions from a large context (five frames to the left and right, respectively). A further refined system was then constructed by applying a maximum likelihood linear transform (MLLT) upon the LDA feature.\footnote{See \url{http://kaldi-asr.org} for more information about this setup.} At this point we have our final GMM acoustic model ready for use in offline decoding. However further processing is necessary to prepare the model for online decoding, using a script provided in Kaldi. This process accumulates statistics for basis-fMLLR computation, computes basis matrices and then accumulates statistics for the online alignment model. This model is then ready for online GMM decoding.

We also train a DNN model using the
{\tt train\_multisplice\_accel2.sh} script provided in Kaldi, which at the time of writing was the recommended script to use for DNN training\footnote{At the time of writing the DNN scripts are under continuous development by the Kaldi team as DNN approaches for speech recognition are a highly active area of research. See the Kaldi website \url{http://kaldi-asr.org} for the latest information about the DNN setup.}. We used four hidden layers and trained over one epoch, which came to 62 iterations. The initial effective learning rate was 0.005 and the final rate was 0.0005.

\subsection{Language models and pronunciation dictionary}
\label{langmodel}

The main focus of Kaldi is on acoustic modelling. There are no language modelling tools provided, but there are some links provided in the usage examples that come with the software. 

For our purposes we require two main types of language models. The first are large language models which encapsulate the language found in the training data. For this we use the MIT n-gram language modelling tool \citep{hsu08} which takes as input a text corpus and generates an n-gram model which can then be input into Kaldi. We combine the text of the PF and WSJCAM corpora as input to this tool and then compile a bigram language model for use in Kaldi, following the Voxforge example in Kaldi.

The second type of language model we require are small constrained grammars, which we use during the multiple choice quiz in the Healthy Living Interaction. For this purpose we used the JSpeech Grammar Format (JSGF)\footnote{\url{http://www.w3.org/TR/jsgf/}} to specify the grammar. We used the {\tt sphinx\_jsgf2fsg} tool included in the {\tt sphinxbase-utils} package\footnote{\url{http://packages.ubuntu.com/precise/sound/sphinxbase-utils}} for the Ubuntu Linux operating system to compile the source JSGF grammar file into a finite-state machine (FSM) representation. Next we use OpenFST tools \citep{allauzen07} to compile the FSM into a finite-state transducer (FST). Finally we used the scripts provided in the Kaldi examples to create the final language model and decoding graph from the FST.

For both types of language model we require a pronunciation dictionary that lists for each word in the dictionary the phonetic sequence(s) for the word. We use the Beep\footnote{\url{ftp://svr-ftp.eng.cam.ac.uk/pub/comp.speech/dictionaries/beep.tar.gz}} dictionary for this purpose, since it is designed for British English pronunciations. For words that are not in the dictionary (e.g. robot names, such as Zeno) we use the Sequitur tool \citep{bisani08} to estimate the phone sequences given the letters of the word. This gives a guess of the pronunciation of the word where it is not available. For small vocabularies it is feasible to manually check these pronunciations to ensure they are correct.

\subsection{Online decoding}

The Kaldi toolkit has been designed primarily for researchers working the field of speech recognition. Most of the framework is based around offline experiments, where the speech audio has been segmented into individual utterances and manually transcribed. The toolkit offers various scripts for training and evaluating with this data.

For our use-case in EASEL (and for robotics applications in general) we require a quite different setup, where we have a live audio input stream and we expect recognition results to be returned as soon as possible at the end of each utterance. In speech research terminology this is referred to as an \emph{online}\footnote{This is perhaps a somewhat confusing term. It has nothing to do with being connected to the internet which is a common contemporary use for the word `online'} decoding system. The decoding algorithms required for online decoding are similar to that used in the offline decoding, except that they are optimised to give a result at the end of the utterance with a very low latency.

An example for online decoding is provided in the toolkit, however this example is quite limited. In order to create a working system for our use case, a significant amount of effort has been invested to refactor and extend this example. We have designed this additional code to be a lightweight extension to Kaldi itself, so that it can be easily installed on top of the Kaldi system. The extra code has been written in C++, the same as Kaldi itself and so can be configured to run very efficiently with little extra overhead. In addition we have designed the code to be modular and re-usable so that other end-users such as roboticists who want to integrate ASR into their own systems can do so with as little effort as possible. 

The extra features implemented are listed below. All features can be easily configured by changing parameters in a shell script.

\begin{itemize}
\item Record audio input from the microphone to file. This code is embedded within the ASR module, meaning that we can record the exact input received by the system, an essential feature for replicating results for experimental analysis. 
\item Choose a data source. The user can switch between live microphone input, and a pre-recorded audio file. When receiving input from the audio file the program runs exactly as it would if it were receiving the data from the microphone; this feature allows us to run the decoder over recorded audio from previous sessions and see how well the recogniser would perform in that situation.
\item Specify an acoustic model. This means we could choose between child and adult acoustic models, or models suited better for different microphones and so on.
\item Dynamic language models. This allows language models to be switched at run-time given an input event from another source. For example in our use-case this allows use to switch between different sets of answer options for each multiple choice question in the quiz. This narrowing of the search space greatly improves recognition accuracy.
\item Specify output options. We have a specific C++ class to deal with outputting the results of the ASR. Currently we write the ASR transcription output to screen and to a log file. This could be expanded in future to allow more detailed information such as n-best paths with associated probabilities.
\item Logging features. All output is logged automatically to date-stamped files for later analysis and debugging.
\end{itemize}

\subsection{Integration with other modules}
\label{integration}

An ASR module is only useful in robotics applications if it is well integrated with the dialogue system and other modules in the wider system. In order to integrate the ASR with the other modules in the EASEL system, we used the YARP middleware \citep{metta06} which is used throughout the EASEL project and has been widely used in many European robotics projects. YARP provides the functionality for programs running across a range of operating systems and programming languages to communicate at run-time via \emph{ports}. Each program can read and write to any number of ports, allowing an easy and flexible form of communication between the modules. Ports can be assigned any name, but in our project we use a convention to have descriptive names to make it easy to identify the nature of the port from the name. For the ASR module, we have developed a simple YARP wrapper that takes the text output from the ASR module and writes it to a YARP port {\tt /SpeechRecognition/Sentence}. This information can then be read by any other module which is interested.

As well as writing out information, the module is also able to receive information at run-time from other modules. Specifically in our use-case we make use of two main types of input events. The first is from the robot itself, via the dialogue manager. These events tell us when the robot has started or finished speaking via its onboard speech synthesiser. Using this information allows us to effective start and stop listening at appropriate times in the dialogue. Since the robots synthesiser voice can be quite loud (especially in a public setting such as museum), we switch off the recognition of human speech while the robot is speaking since otherwise we would end up recognising the robot speech instead of the human speech. The audio is still recorded to file (for later analysis), but it is not input into the recogniser. This is a simple but effective way to filter out the robot speech - although evaluations showed it is not perfect (see discussion in Section \ref{segmentation_results}).

The second type of input event is the specification of language models, mentioned briefly in the previous section. In our use-case this event is determined by the dialogue manager, which keeps track of which part of the dialogue the system is currently engaged in. In the first part of the interaction the robot asks the participant to repeat certain keyphrases to allow for voice adaptation. Thus the language model for this stage is very simple; a grammar model that has only two possible paths - either the entire keyphrase or silence/non-speech. Later on in the interaction, the grammar is slightly more complex - the robot presents a question or instruction with multiple choices (answers to the question, or a selection of commands to give to the robot); for this part of the interaction the grammar comprises the set of choices, again plus an option for silence/non-speech if nothing is uttered. Since in our use-case we know the expected utterances for each stage in the dialogue we precompiled the grammars in advance. Then during the interaction the dialogue manager can send a simple message to the ASR module to specify which grammar should be used for the next utterance. This information greatly improves the recognition performance since we have a much smaller search space than if we had to include all possible utterances in a single language model.

\section{Experiments with existing corpora}
\label{offline_evaluation}

We evaluated our Kaldi ASR setup with existing corpora. We used the GMM and DNN acoustic modelling and decoding approaches described in Section \ref{acousticModelling}. We used a bigram language model computed using the MITLM toolkit as described in Section \ref{langmodel}, using all the text in the corpora to compute the bigrams. We use the WSJCAM (adult) and PF (child) corpora described in Section \ref{trainingData}. We add various levels of noise to the test data (as described in Section \ref{noiseAugmentation}) to evaluate how well the algorithms perform under these conditions.

First, we report the results obtained using only clean training data in Table \ref{cleanTab}.

\begin{table}[htp]
\begin{center}
\begin{tabular}{|c|c|c|c|c|c|} \hline
Algorithm & Test set &  Clean & 20dB & 10dB & 5dB \\ \hline
GMM & WSJ &  7.9 & 12.7 & 45.6 & 79.5 \\ \hline
GMM & PF & 15.7 & 30.2 & 54.0 & 64.8 \\ \hline
DNN & WSJ & 4.9 & 12.3 & 47.7 & 78.1 \\ \hline
DNN & PF & 15.6 & 30.7 & 55.1 & 65.2 \\ \hline
\end{tabular}
\caption{WER (\%) using clean training data. For pure noise the GMM WER was 0.7 and for the DNN WER was 2.3.}
\label{cleanTab}
\end{center}
\end{table}

The DNN approach significantly outperforms the GMM approach on the clean adult (WSJ) speech with a WER of 4.9\% for the DNN approach and 7.9\% for the GMM approach. Interestingly this performance is not replicated for the child (PF) speech, with performance roughly similar in both cases (this maybe due to the parameter settings; it is possible that adjusting parameters such as the number of hidden layers, the learning rate and so on may improve results, but these were not investigated in depth for this work).

As expected, adding noise to the data causes performance to degrade; with only a 20dB signal to noise ratio the performance for the GMM and DNN approaches is roughly levelled, both obtaining 12 to 13\% WERs for adult speech, and 30\% for the child speech. The degradation in performance continues with higher levels of noise.

Next we report results using the full set of training data including all the noise-augmented data in Table \ref{noisyTab}.

\begin{table}[htp]
\begin{center}
\begin{tabular}{|c|c|c|c|c|c|} \hline
Algorithm & Test set &  Clean & 20dB & 10dB & 5dB \\ \hline
GMM & WSJ &  11.5 & 10.0 & 21.1 & 42.9 \\ \hline
GMM & PF & 23.4 & 24.2 & 41.1 & 54.0 \\ \hline
DNN & WSJ & 6.5 & 5.4 & 9.7 & 24.0 \\ \hline
DNN & PF & 9.9 & 13.1 & 27.4 & 43.3 \\ \hline
\end{tabular}
\caption{WER (\%) using full training data including all noise levels. For the noise data, the GMM WER was 5.04 and the DNN WER was 1.16.}
\label{noisyTab}
\end{center}
\end{table}

In this case the GMM approach fares poorly. The results for the clean test data are worse, suggesting that the models are unable to discriminate between noise and speech. However the results for the noisy data are better than before. Interestingly the performance for adult speech with 20dB noise level is slightly better than for clean speech.

However the DNN approach performs much better. The performance on clean speech is similar to before, and in fact the performance on clean child speech is better, 9.9\% vs 15.6\%. This surprising result could be caused by the DNN configuration; it is possible that the configuration was better suited for the larger training set size, and could be improved for the smaller set with some tuning of parameters such as learning rate. The results for noisy data are also superior compared to the GMM approach. For adult speech the WER is kept below 10\% even with 10dB noise. Good performance is also achieved with the child speech, although the results suggest that child speech is harder to discriminate against background noise. However the DNN approach proves to be far more robust than the GMM approach in the presence of noise.

\section{Live experiments}
\label{live_evaluation}

\begin{figure}[htp]
\centering
\includegraphics[width=130mm]{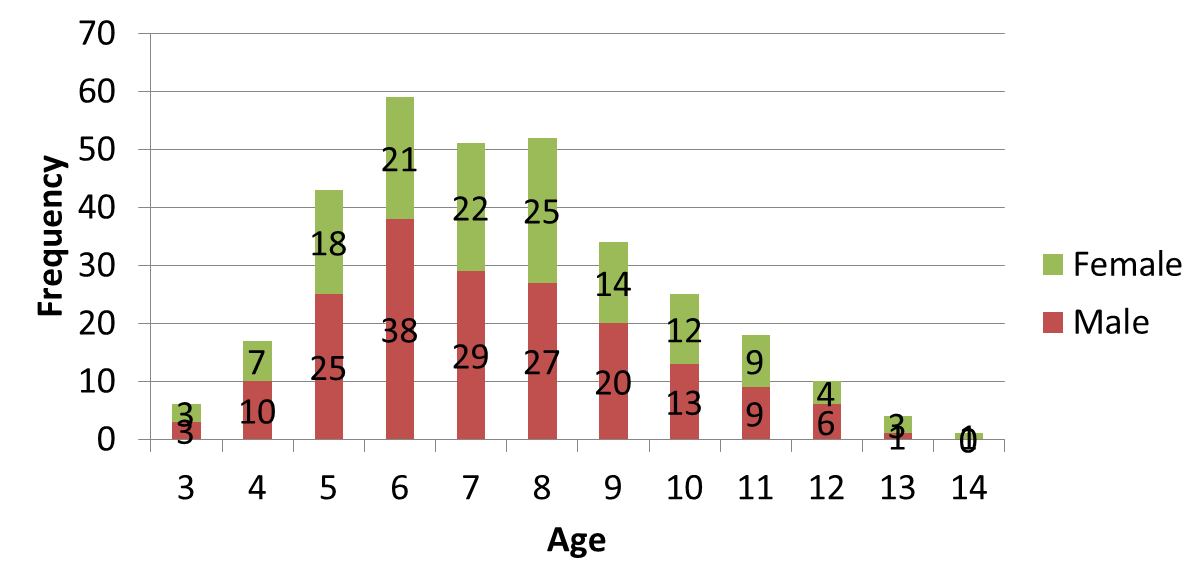}
\caption{Distribution of participants by age and gender.}
\label{age_gender_dist}
\end{figure}

We installed the system at the Weston Park museum in Sheffield, as described in Section \ref{healthyLivingInteraction}. The event ran for 14 days during the Easter holidays. In total we had 367 individuals\footnote{This figure does not include a number of individuals who took part in the interaction but were not recorded. Some children wanted to try but for whatever reason did not want, or were not able, to consent to participate in the study (for example because their parents were not present, or because on one occasion we ran out of paper consent forms). The figure also does not include some adults who participated, since our focus was on child speech.} who we noted as participating in the interaction. Of these, we had 329 interactions where parents/guardians provided consent to participate in the study. From this subset, we had 326 who gave consent for audio to be recorded during the interaction, and 321 who provided consent for video recording. Of the 326 children for which we had consent for audio recording, we exclude one case where we did not record the age, and five cases where the adult spoke for the child, leaving a total of 320 participants for which we have the full set of data. This set of 320 comprises 181 male (57\%) and 139 female (43\%) children. The detailed age gender distribution of the 320 participants is shown in Figure \ref{age_gender_dist}. 

\begin{figure}[htp]
\centering
\includegraphics[width=100mm]{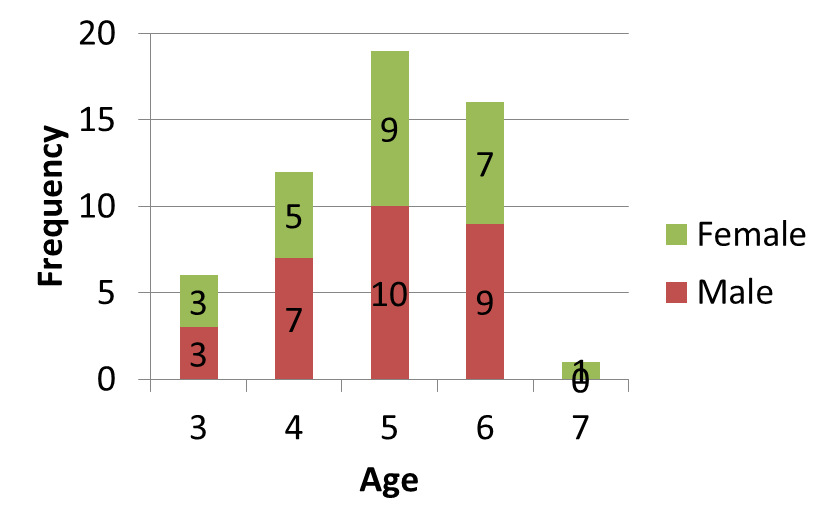}
\caption{Distribution of wizarded interactions by age and gender.}
\label{woz_dist}
\end{figure}

Of these 320 participants, we wizarded in 54 cases. The decision to wizard the interaction was based on the age of the child; the level of background noise in the room; and a quick judgement after initial conversation with the child to determine if they would be able to speak fluently with the robot. The distribution of wizarded interactions is shown in Figure \ref{woz_dist}.

\subsection{Data transcription}

All the speech audio files we obtained were manually transcribed. Transcribers used the Xtrans software\footnote{\url{https://www.ldc.upenn.edu/language-resources/tools/xtrans}} created by the Linguistic Data Consortium to annotate the files. The transcribers were provided with the automatically generated transcriptions that were created during the running of the experiment. They were asked to check each of these and correct as necessary, and also to search for other utterances that may have not been detected by the ASR. They were instructed that the segments of interest were clear speech from the child participant that was directed to the robot; they were asked to exclude noise from the robot, speech or noise from other people in the room, and speech from the child to others in the room, especially if it was whispered or muttered. Since we wanted to test the effectiveness of the ASR in background noise we asked the transcribers to leave such noise at the beginning and end of the speech segments; this enables us to test if our ASR module can effectively distinguish the intended speech from the background noise. 


In some cases the automatic segmentation of the utterances was incorrect (see discussion in Section \ref{segmentation_results}); transcribers were asked to adjust the start and endpoints of the automatic segments to ensure that the human speech was included, and the robot speech excluded. However, as noted above, we asked transcribers to leave background noise at the start and end of the auto-generated segments, and only adjust the segment if the human speech had been missed, or if the robot speech had been erroneously included.

\subsection{Fluent and in-vocabulary speech}

From the transcriptions we are able to readily classify utterances into fluent and disfluent utterances. Disfluent or mispronounced speech was marked with symbols to indicate mispronunciations and false starts. We marked all such utterances as disfluent, and all other utterances as fluent.

We further divide the fluent utterances into expected and unexpected utterances based on whether or not the utterance appears in the recognition vocabularies. Altogether there are only 24 phrases that are expected during the interaction (e.g. `Hello Zeeno I am ready to start', `Moved slowly for ten seconds' etc.) We also included the tag `!SIL' as an expected utterance; this tag denotes non-speech/background noise. All other (fluent) utterances were marked as unexpected. 

\begin{figure}[htp!]
\centering
\includegraphics[width=130mm]{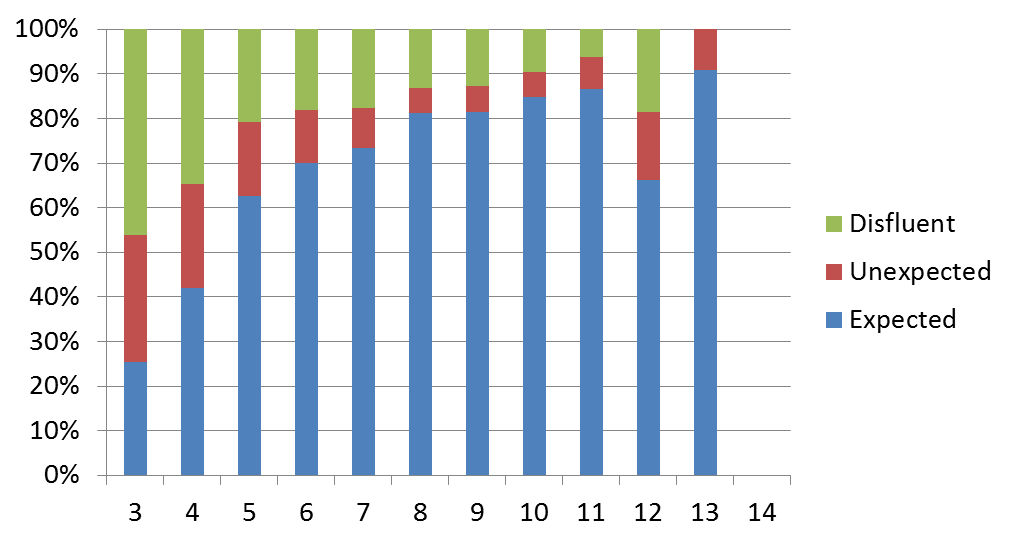}
\caption{Distribution of fluent/expected speech for male children, age on x-axis.}
\label{fluency_male}
\end{figure}

\begin{figure}[htp!]
\centering
\includegraphics[width=130mm]{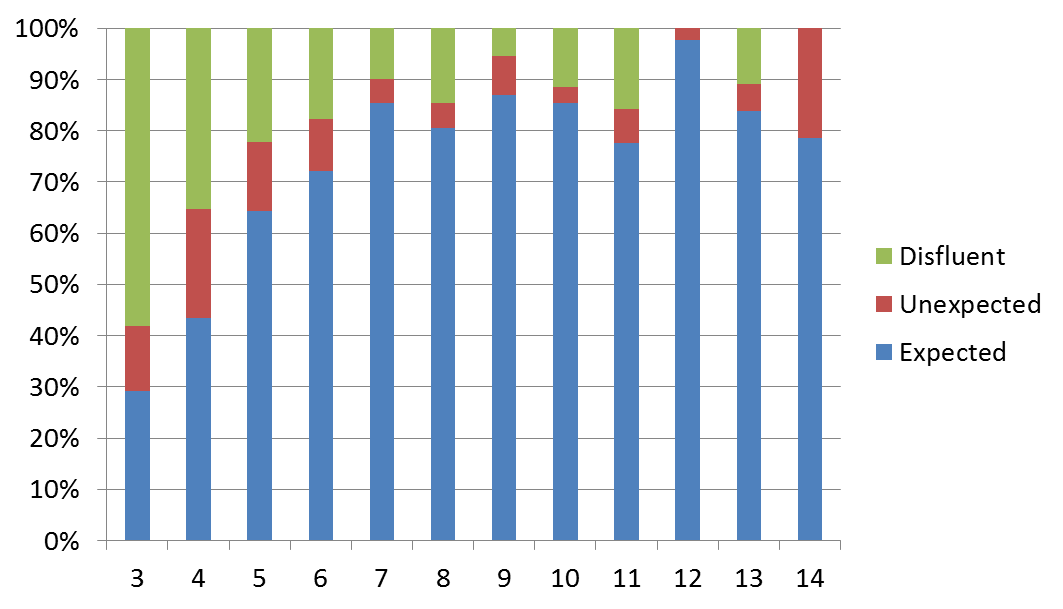}
\caption{Distribution of fluent/expected speech for female children, age on x-axis.}
\label{fluency_female}
\end{figure}

In total we had 4707 transcribed utterances. Of these 854 were marked as disfluent while 3853 were marked as fluent. Of the fluent utterances, 3351 were in the expected set, while 502 were not. The distribution of fluent and expected speech by age is shown for male children in Figure \ref{fluency_male} and for female children in Figure \ref{fluency_female}.

\subsection{Accuracy of recognition during live experiments}

In this section we evaluate the accuracy of the ASR module during the live experiments. For this purpose we focus on the fluent, expected utterances as described in the previous section.

We ran with ASR for 266 of the 320 participants, however there was a system crash for one participant leaving a test set of 265. For these 265 participants, we obtained a total of 3596 utterances. Of these we found 3087 to be fluent, while the remaining 509 to contain some disfluency. Of the fluent utterances we found 2770 to be expected (i.e. in-vocabulary) utterances. This set of 2770 provides the evaluation set for which we examine the live ASR performance; we would expect all these utterances to be correctly recognised by the ASR.

To evaluate the accuracy of the ASR we compared the logged output of the ASR against the manually transcribed gold standard. For each of the 2770 utterances we checked if the logs showed the utterance had been correctly recognised by the ASR. Since the manual transcription allowed for the start and end-times of segments to be altered slightly to fit the utterance, we allowed for this in the evaluation, checking segments in the logs which overlapped the gold-standard segments. In total we found that 2582 utterances were correctly identified, while 188 were not; giving a classification accuracy of 93.2\%. 


This result includes the utterances at the start of the interaction which are used for adaptation purposes (e.g. `Testing A B C') for which the vocabulary is limited only to that utterance or to silent. This gives a slightly distorted picture since we would expect the recognition results for this set to be higher than for the multiple choice stage where the grammar allows for four options plus silence (although even for the adaptation stage there were occasionally errors where the phrases were not recognised correctly.)

\begin{figure}[htp!]
\centering
\includegraphics[width=130mm]{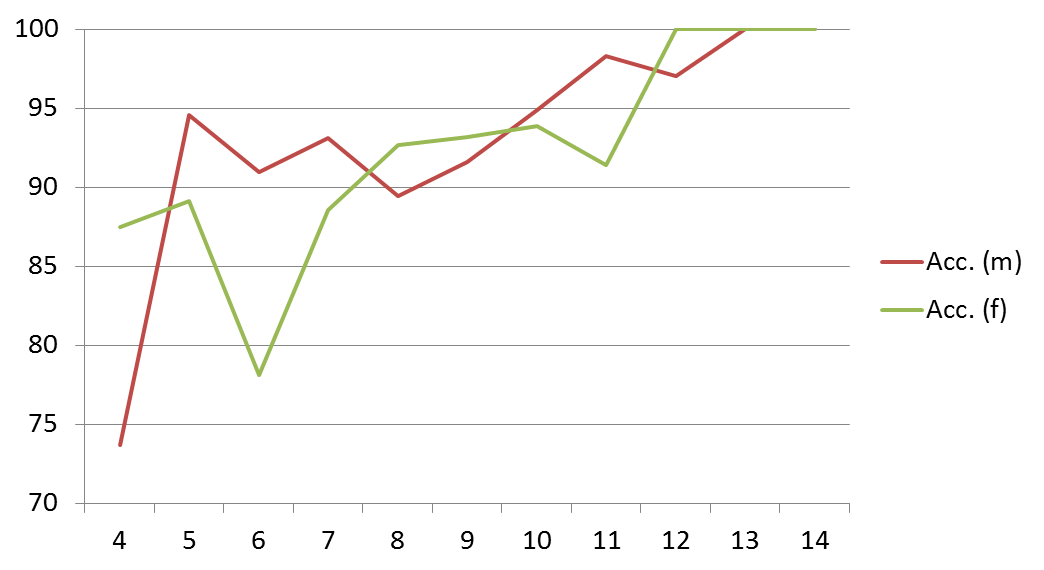}
\caption{Accuracy for multiple choice questions.}
\label{multi_asr}
\end{figure}

We therefore give results excluding this initial set, and also excluding the phrase `Zeeno Start the Quiz' which was also recognised with a single item grammar, leaving only the multiple choice phrases. This gives a total of 1771 utterances for consideration; of these 1616 were correctly identified, and 155 were not. The accuracy over the multiple choice stage is therefore 91.2\%. The accuracy for age and gender classes is shown in Figure \ref{multi_asr}.

The above figures include only fluent, in-vocabulary speech. We also evaluated performance on speech with minor disfluencies. These were found by inspecting the transcriptions for utterances which contained more than 75\% of the words of an accepted utterance (limiting only to multiple choice answers). This allowed for minor disfluencies or false starts. In total 184 utterances were found to fit this criteria, and were evaluated in the same way as before, with 136 found to be correctly identified, and 48 not, giving an accuracy of 73.9\%. When these minor disfluent sentences were included with the fluent, in-vocabulary speech, the total was 1955 with 1752 identified correctly, and 203 not, giving an overall accuracy of 89.6\%. This confirms that even with minor disfluencies the system gives good results. 


\subsection{Segmentation errors}
\label{segmentation_results}

As described in Section \ref{integration}, the ASR module received information from the robot when it was about to start speaking, and when it had finished speaking. This information was provided by the Acapela speech synthesiser onboard the robot. In principle, these messages should have been enough to ensure that the ASR could switch off while the robot was speaking, and the re-activate as soon as it had finished. However this did not work as planned, since the end-of-speech message would often come late from the robot (usually the delay was about half a second). This would then result in the ASR being activated too late, meaning that the first part of the human speech utterance was missed. 

In order to adjust for this problem, we applied a simple fix. On receipt of the end-of-robot-speech message, the ASR module would read back about half a second, to account for the expected delay. In many cases, this method worked well, providing the start of the human utterance on time, where it would otherwise have been missed. However in some cases the method did not work as intended. In some cases the module would read back too far (in which case the ASR module would receive the last part of the robot speech along with the utterance). In other cases it would not read back far enough; the delay of the end-of-speech was larger than half a second, so the human speech was still missed at the start.

If there was such a problem with the start point of the segmentation during the live interaction, it would affect the accuracy of the ASR in two ways. First, it would affect the recognition of that utterance itself, since it would either have missed the start of the utterance, or it would have included extra audio at the start from the robot speech. Secondly, it could also affect accuracy for subsequent utterances, since the ASR would run feature adaptations based on the recognised utterances. If the utterance had been badly segmented then the adaptation would be performed with the wrong input data, creating a malformed adaptation.

The endpoints of the automatic segments were also sometimes incorrect. This did not cause a problem during the live interactions, but does cause a problem for the offline analysis done afterwards. In some cases the endpoint occurs too early; this happens when the ASR has successfully recognised the utterance before the end of the utterance; it would then immediately mark this as an endpoint. In other cases the endpoint is late; this occurs where the ASR did not successfully recognise the utterance, resulting in the dialogue manager timing out, and the robot starting to speak.

From the manual transcriptions, we were able to determine the frequency of each type of segmentation error. For each of the 2791 fluent, in-vocabulary utterances in the gold standard, we found the best matching utterance in the automatic transcriptions. This was done using the following approach; first we tried to find an overlapping segment marked with the same utterance transcription as the gold standard; this was then marked as the best matching utterance. If this could not be found, then we instead looked for the overlapping segment with the minimum distance from the gold standard segment. Once each gold standard segmented utterance was matched to the corresponding segment in the automatic set, we were then able to compare to find segmentation errors. We classified these into early starts, late starts, early ends and late ends, shown in Table \ref{segmentation:tab}.

\begin{table}[htp]
\begin{center}
\begin{tabular}{|c|c|} \hline
Error & Frequency  \\ \hline
Early start & 398 \\ \hline
Late start & 77 \\ \hline
Early end & 116 \\ \hline
Late end & 115 \\ \hline
\end{tabular}
\caption{Frequency of segmentation errors in the 2791 fluent, in-vocabulary utterances.}
\label{segmentation:tab}
\end{center}
\end{table}

The most common segmentation error by far was the `early start' error; this shows that the quick-fix method of reading back in the data was reading back too far, including snippets of the end of the robot speech. However there were still a number of `late start' errors; this means that sometimes that end-of-speech was coming in so late that even the fix was not enough to account for it.

The `early end' errors can be interpreted as a success for the ASR (provided the recognition was correct); the latency is so low, that the system is able to recognise the utterance \emph{before} the person has finished uttering it. This low latency is very useful for robotic applications where fast dialogues may be required. The `late end' conversely usually indicates a failure of the ASR to recognise the utterance.

\section{Conclusions and Further Work}
\label{conclusions}

In this article we presented our work on developing an ASR module for use in the EASEL project. Our module has been integrated into a multimodal system using a humanoid robot and vision sensing. Using this system we have developed the `Healthy Living Interaction' which helps children learn about exercise and energy in an interactive way. This system has been successfully deployed in a public museum setting, with 325 children participating in the interaction for research purposes. We have presented an initial analysis of the collected data in this article. We have measured the number of fluent and expected utterances obtained and how these have varied depending on gender and age. We have also evaluated the accuracy of the ASR module over these interactions, with results showing that over 90\% of fluent, in-vocabulary utterances have been correctly recognised.

In future work, we plan to utilise this valuable data set for further improving our system. We will use the labelled data as training and test data for thorough evaluations. We will work on improving the segmentation method, using a more reliable approach that finds the end of robot speech from the audio signal itself, rather than relying on messages to be transmitted from the robot. 

We would like to make the source code of the system publicly available, and we invite other researchers in related fields of robotics, speech and language processing and more to collaborate to further develop this system.

\section*{Acknowledgements}

This work is supported by the European Union Seventh Framework Programme (FP7-ICT-2013-10) under grant agreement no. 611971.

We would like to thank Theo Botsford and Jenny Harding for their help in developing the interaction setup and running the event at the museum, and also Jon Barker and Saeid Mokaram for their technical advice in developing the ASR. 





\section*{References}

\bibliographystyle{elsarticle-harv}
\bibliography{mybib}





\end{document}